\documentclass[10pt, conference, letterpaper]{IEEEtran}

\usepackage{cite}
\usepackage{subfigure}
\usepackage{url}
\usepackage{verbatim,epsf}
\usepackage[caption=false,font=footnotesize]{subfig}
\usepackage{enumitem}
\usepackage{amssymb}
\usepackage[noend]{algpseudocode}
\usepackage{algorithmicx}
\usepackage{tabularx}
\usepackage{balance} 
\usepackage{amssymb}
\usepackage{amsmath}
\usepackage{graphicx}
\usepackage{epstopdf}
\usepackage{epsfig, color}
\usepackage{comment} 
\usepackage{amsfonts} 
\usepackage{amsmath} 
\usepackage{rotating}
\usepackage{amssymb}
\usepackage{amsmath}
\usepackage{amsfonts}
\usepackage{wasysym}

\usepackage{algorithm}
\usepackage{color}

\usepackage{verbatim} 
 
\newsavebox{\ieeealgbox}

\newlength\myindent
\IEEEaftertitletext{\vspace{-0.28in}}

\begin{document}	
\IEEEoverridecommandlockouts

\title{\huge
	{LIDAUS: Localization of IoT Device via Anchor UAV SLAM}}
\author{
	\IEEEauthorblockN{Yue Sun\IEEEauthorrefmark{2}, Deqiang Xu\IEEEauthorrefmark{1}, Zhuoming Huang\IEEEauthorrefmark{1}, 
		Honggang Zhang\IEEEauthorrefmark{1}, Xiaohui Liang\IEEEauthorrefmark{2}\\
		\IEEEauthorrefmark{1}Engineering Dept., UMass Boston, Boston, MA. 
		\IEEEauthorrefmark{2}CS Dept., UMass Boston, Boston, MA.\\
		Email: \{yue.sun001, deqiang.xu001, zhuoming.huang001, 
	honggang.zhang, xiaohui.liang\}@umb.edu}
}

\maketitle
\IEEEpubidadjcol

\pagestyle{plain}

\begin{abstract}
We introduce LIDAUS (\textbf{L}ocalization of \textbf{I}oT \textbf{D}evice via 
\textbf{A}nchor \textbf{U}AV \textbf{S}LAM), an infrastructure-free, multi-stage SLAM system
that utilizes an Unmanned Aerial Vehicle (UAV) to accurately localize IoT devices
in a 3D indoor space where GPS signals are unavailable or weak, 
e.g., manufacturing factories, disaster sites, or smart buildings. 
The lack of GPS signals and infrastructure support makes most of the existing indoor localization systems 
not practical when localizing a large number of wireless IoT devices.
In addition, safety concerns, access restriction, and simply the huge amount of IoT devices
make it not practical for humans to manually localize and track IoT devices.
To address these challenges, the UAV in our LIDAUS system 
conducts multi-stage 3D SLAM trips to localize devices 
based only on RSSIs, the most widely available measurement of the signals of almost all commodity IoT devices.
The main novelties of the system include a weighted entropy-based clustering algorithm to select high quality RSSI 
observation locations, a 3D U-SLAM algorithm that is enhanced by deploying anchor beacons along the UAV's path, 
and the path planning based on Eulerian cycles on multi-layer grid graphs that model the space in exploring stage and 
Steiner tree paths in searching stages. 
Our simulations and experiments of Bluetooth IoT devices have demonstrated 
that the system can achieve high localization accuracy based only on RSSIs of commodity IoT devices.

\end{abstract}

\section{Introduction}

The rapidly increasing deployment of Internet of Things (IoT) devices around the world 
is changing many aspects of our society. 
IoT devices can be deployed in various places for different purposes, 
e.g., in a manufacturing site or a large warehouse, and they can be displaced over time due to human activities.
Even though they make our lives better, 
it is quite challenging to keep track of them in a GPS weak or no GPS space (e.g., an indoor space)
due to their pervasive presence, deployment scale, 
access constraints/restriction, and even safety concerns (e.g., disaster sites or sites with 
harmful radiation).

In recent years, indoor localization of wireless devices or sensor nodes in general has been an active research area
\cite{bulten2016human, an2020mobisys, vasisht2016decimeter,sen2012you, soltanaghaei2018multipath, lanzisera2011radio, xiong2014synchronicity, youssef2006pinpoint, shu2015gradient, ayyalasomayajula2018bloc, kotaru2015spotfi,
altini2010bluetooth, wu2012fila, werb1998designing, kumar2014single, anchor2007sensor}.
For example, \cite{bulten2016human} applied 
Received Signal Strength Indicator (RSSI) in localizing smart devices. 
However directly utilizing RSSIs can result in large estimation errors 
due to shadow fading and multi-path fading in complex indoor environments \cite{yang2013rssi}.
To address the problem, research has been done on improving RSSI-based techniques, e.g., via better 
fingerprinting \cite{shu2015gradient,altini2010bluetooth}. 
Furthermore, Channel State Information (CSI) has been utilized 
to get more accurate localization, 
but usually these methods need to build customized equipment and/or require some coordination between wireless devices in a space \cite{wu2012fila, soltanaghaei2018multipath, lanzisera2011radio, xiong2014synchronicity,
	youssef2006pinpoint, werb1998designing, kumar2014single}. 
The recent trend of applying neural networks to indoor localization relies 
on a pre-built model based on extensive data collection and model training \cite{an2020mobisys}\cite{ayyalasomayajula2020deep}, 
which is not applicable to a dynamic changing space. 
Furthermore as of today, CSI information cannot be easily obtained from Bluetooth signals (unlike RSSI) even though 
some recent research \cite{ayyalasomayajula2018bloc} has made some effort in this direction. 
Since Bluetooth has been the dominating technology in IoT device market and 
its global market size is expected to reach $\$58.7$ billion by 2025 \cite{bluetooth},
a universal IoT device localization system should certainly include Bluetooth devices. 

Our goal is to develop an infrastructure-free IoT device localization system 
based only on RSSI, due to its ubiquitous presence and wide availability in the existing commodity IoT devices,
especially Bluetooth devices. We aim to design a system that  
(1) can work in an indoor space where GPS is not available or a space that is not accessible for humans,
(2) does not require any fingerprinting or pre-trained model and can deal with dynamic changing environment, 
(3) is easily deployable (without any customized signal processing hardware).
IoT devices are also referred to as \textit{beacons} in this paper. 

We develop LIDAUS (\textbf{L}ocalization of \textbf{I}oT \textbf{D}evice via 
\textbf{A}nchor \textbf{U}AV \textbf{S}LAM), an infrastructure-free, multi-stage 
SLAM\footnote{Simultaneous Localization and Mapping (SLAM) 
	refers to a class of techniques used by a mobile robot to explore an environment with unknown landmarks, with 
	a goal to develop a map of the environment (i.e., to localize the landmarks) 
	and in the mean time to localize itself within the environment during its trip.}
system that utilizes a UAV (Unmanned Aerial Vehicle) to search and accurately localize IoT devices
in a space. 
The major contributions of this paper are summarized as follows.
\begin{itemize}
	\item We design and evaluate LIDAUS, 
	which consists of an \textit{exploring stage} that derives an initial estimation of the positions of 
	the IoT devices (referred to as \textit{target beacons})
	in a 3D space, and multi-round \textit{searching stages} to accurately localize all devices.
	It discretizes a 3D space into multiple horizontal layers, with each layer being modeled as
	a grid graph. The exploring stage adopts a UAV path planning based on an Eulerian cycle
	that covers all edges of the grid graph of each layer, in order to resolve 
	direction or angle ambiguity that is inherent in RSSI-only distance estimation.
	The UAV's path in each searching stage is based on a Steiner tree that allows the UAV
	to approach each beacon's estimated position with minimum cost of deploying anchor beacons,
	which are dynamically deployed by the UAV along its path 
	to mitigate the impact of the unreliable noisy RSSIs of target beacons in its SLAM computation.

	\item 	
	We design \textit{U-SLAM}, a SLAM algorithm based on FastSLAM\footnote{
		Unless specified otherwise, FastSLAM means FastSLAM 1.0 in this paper. We find that FastSLAM 2.0 performs 
		not as well as FastSLAM 1.0 as RSSIs are highly unreliable. } \cite{fastslam2003thesis}
	that can do SLAM in a 3D search space. 
	We introduce a \textit{weighted entropy-based clustering algorithm} 
	that selects a cluster of positions where the RSSIs observed for a specific target beacon
	can be used in the location estimation of the beacon in an offline \textit{U-SLAM replay} 
	to improve the beacon's localization accuracy.

\end{itemize}	

LIDAUS has the following advantages: 
The system is easily deployable, and cost effective as it does not require an infrastructure with  
some specialized hardware (e.g., customized designed antenna arrays in lots of existing work).
It can deal with a dynamically changing environment as no pre-trained models are needed.  
It does not require any data communication or computation from target beacons, i.e., 
the system is completely passive. 
It can be used in a broad range of environments such as 
complex indoor space where GPS signals are not available, or a place where it is risky or even impossible for humans to operate, 
e.g., a disaster or a high radiation site.

LIDAUS is applicable to various IoT devices with RSSIs. In this paper, 
we focus on Bluetooth RSSIs 
and have built a prototype system by using the open source drone platform \cite{crazyflie} to evaluate 
the performance of the system.
Through simulations and experiments, we have demonstrated the effectiveness of our system 
in achieving high localization accuracy, with an average localization error less than $1m$. 
LIDAUS also significantly outperforms a random SLAM fly method 
and a naive SLAM search method, with $258\%$ and $336\%$ accuracy improvement respectively.
Our study illustrates an important direction 
to improve the accuracy of IoT device localization with low cost and high efficiency.

The rest of the paper is organized as follows. 
Section \ref{sec_related} discusses related work. 
Section \ref{sec_background} presents the background and the challenges of RSSI-based localization. 
Section \ref{sec_system} discusses the design of our system. 
The performance evaluation of the system is given in Section \ref{sec_eval}. 
Finally the paper concludes with Section \ref{sec_conc}.
 
\section{Related Work}\label{sec_related}

Indoor localization of wireless sensors, devices, or humans has been an active research topic 
in recent years \cite{vasisht2016decimeter,sen2012you, soltanaghaei2018multipath, lanzisera2011radio, xiong2014synchronicity, youssef2006pinpoint, shu2015gradient, ayyalasomayajula2018bloc, kotaru2015spotfi,
	altini2010bluetooth, wu2012fila, werb1998designing, kumar2014single, anchor2007sensor}.
RSSI has been utilized in many range-based indoor localization 
system, but RSSI exhibits high temporal and spatial variance due to the multipath effect. 
To address this issue, research has been done on improving RSSI-based techniques, 
e.g., a gradient-based fingerprinting method in \cite{shu2015gradient} and 
a neural-network based approach in \cite{altini2010bluetooth}. 
In addition, CSI-based method has also been studied extensively, e.g., \cite{wu2012fila}.
 
One of the most recent work on localizing IoT devices is iArk \cite{an2020mobisys}. 
But unlike our work, iArk needs to design a large customized antenna array with pre-trained neural network models, 
which makes iArk not applicable for a dynamically changing environment. 
Another related work, HumanSLAM \cite{bulten2016human}, also utilizes SLAM based on the RSSIs received 
by smartphones carried by human users walking around
to locate IoT devices. Different from \cite{bulten2016human}, 
our system can be applied in an environment which is dangerous for human to access, 
e.g., a nuclear plant site or a disaster site.

Similar to our idea of using anchor beacons, 
\cite{anchor2007sensor} proposes a method to let sensor nodes to compute 
their own positions based on the received signals from a set of pre-deployed and position-known anchor nodes. 
But in our work, anchor nodes are dynamically deployed with only estimated positions in a UAV search process.  
Our work is also related to the rich literature in robotics on applying mobile robots in search and rescue.
FastSLAM \cite{fastslam2003thesis} is adopted as an important component in our system.
Application of UAVs in various mobile scenarios has also been active research area. For example, \cite{piao2019automating}
introduces a method that lets a UAV to automatically collect CSI measurements for finger-printing based localization. 
\cite{motlagh2017uav} discusses a vision-based UAV system for crowd surveillance.

\section{Background and Challenges}\label{sec_background}
 
In this section we discuss the background of RSSI-based distance estimation, 
the challenges of applying it to wireless IoT device localization,
and some of the design ideas of our proposed system to address those challenges. 

\subsection{Distance estimation based on RSSIs}

The functional relationship between the distance (between a signal transmitter and a receiver) 
and the RSSI measured by a receiver 
is given by the following equation \cite{seidel1992914}:
$RSSI=-10 \alpha \log_{10} d/d_0 +\beta$,
where $\alpha$ is the signal propagation exponent, 
$\beta$ is a reference $RSSI$ value at $d_0$, 
and $d$ is the distance between a signal transmitter and a receiver. 
We usually set $d_0$ to be $1m$ so that we can get the value of $\beta$, 
which is $RSSI$ measured at a distance of $1m$ from the node. 
Then the above equation can be simplified as:\\
\begin{equation}
RSSI=-10 \alpha \log_{10} d+\beta 
\label{eqn_rssi}
\end{equation}

The parameters $\alpha$ and $\beta$ in Eqn. (\ref{eqn_rssi}) are different for different wireless transmitters, and they can be estimated in practice.  We can use Least Square Method (LSM) \cite{everitt2005least} to get the estimated values of $\alpha$ and $\beta$ based on experimentally collected RSSIs and distance values. Specifically we can place a signal receiver at various distances away from a transmitter and measure the $RSSI$s of received signals, denoted by $RSSI_{m}$. Let $d_m$ denote a measured distance between the signal transmitter at position ($x_t, y_t, z_t$) and the receiver at ($x_r, y_r, z_r$).
Based on Eqn. (\ref{eqn_rssi}), we can get an expected value $RSSI_{e}$ as a function of 
a measured distance $d_m$:
$RSSI_e=-10 \alpha \log_{10} d_m +\beta$, where $d_m=\sqrt{(x_r-x_t)^2+(y_r-y_t)^2+(z_r-z_t)^2}$.
Then the estimated parameters $\hat{\alpha}$ and $\hat{\beta}$ can be obtained 
by using $N$ measured pairs $(d_m, RSSI_m)$ in solving the following optimization problem:
\begin{eqnarray}
\begin{aligned}
\hat{\alpha},\hat{\beta}
&=\mathop{\arg\min}_{\alpha,\beta} \sum_{i=1}^{N}(RSSI_{m}-(-10 \alpha \log_{10} d_m +\beta))^2
\end{aligned}
\label{eqn_lsm}
\end{eqnarray}
In this work we use the LSM to derive estimated $\alpha$ and $\beta$.

\subsection{Challenges}

There are two major design challenges of applying the above RSSI-based distance estimation 
in the localization of a wireless IoT device. 
\textbf{First}, the RSSIs of IoT devices are highly variable \cite{yang2013rssi}, 
which is in sharp contrast to the reliable and precise sensor data (e.g., LIDAR data)
used in the conventional SLAM in robotics.
Therefore in a SLAM-based localization system (which we intend to design),
directly feeding RSSI data to a robot's SLAM algorithm 
can result in large estimation errors of both the robot itself and the IoT device to be localized.
\textbf{Second}, RSSIs alone can only be used to estimate the distance between a robot or UAV 
from an IoT device, but not the direction or angle of the device relative to the robot. 
For example, if a device is on the north side of a UAV, 
the UAV flies on a straight line from west to east cannot decide whether the device is on its north side or south
side. This ambiguity is referred to as \textit{angle or direction ambiguity}.

\subsection{Illustration of some key design ideas of our system}

To address these challenges, our LIDAUS system introduces a few novel design ideas.
We now illustrate two of them through simple examples. 
The first one is that after each SLAM trip of a UAV, for a particular beacon, 
the UAV conducts a \textit{SLAM replay} based only on a selected 
set of observation location points that gives high quality RSSIs for the beacon
(discussed in Section \ref{sec_clustering}). 
The second idea is to utilize \textit{anchor beacons} in a SLAM process. 
We let the UAV deploy anchor beacons\footnote{Deploying or releasing a  
	low-cost coin-sized Bluetooth beacon can be done through a simple mechanism 
	attached to the UAV. In addition, a low-cost NRF51822 beacon only costs around \$$4$ \cite{aliexpress}.
	} on its path to 
help improve the accuracy of localizing itself during its trip, 
described later in Algorithm \ref{alg_search_stage} in Section \ref{sec_target_search}. 

\subsubsection{Replay SLAM based on high quality RSSIs}
\begin{figure}[htb!]
	\centerline{
		\begin{minipage}{3.5in}
			\begin{center}
				\setlength{\epsfxsize}{3.5in}
				\epsffile{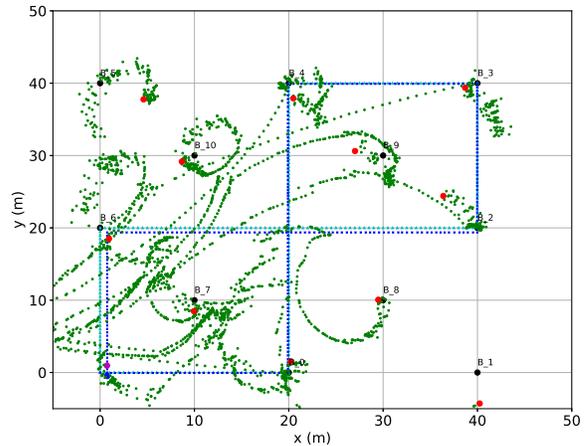}\\
				{}
			\end{center}
		\end{minipage}
	}
	\caption{There are $11$ target beacons: $B\_0, B\_1, ..., B\_10$. Black dots show the actual locations
		of those target beacons. Red dots show the estimated positions of the targets when the drone returns
		its starting position. Green dots show the estimated positions of the targets by the drone in its SLAM process. 
		Blue color dotted lines show the estimated locations of the UAV itself during its trip.}
	\label{fig_targets_multiple}
\end{figure}

We now use a simple 2D example to illustrate this idea. 
Consider $11$ target beacons placed on the floor of a room. 
They are shown as black dots in Fig. \ref{fig_targets_multiple}.

The UAV first does a FastSLAM in a pre-determined path:
$(0,0) \to (20, 0) \to (20, 40) \to (40, 40)
\to (40,20) \to (0, 20) \to (0,0)$ in the room. This initial trip 
gives us an estimation of the locations of those target beacons. 
The blue lines in Fig. \ref{fig_targets_multiple} shows the UAV's path. 
The red dots show the final estimated positions of the beacons, and the green dots show the changing estimates 
during the trip. We see that many beacons cannot be accurately localized 
as indicated by the distance between the black dots and their corresponding red dots in Fig. \ref{fig_targets_multiple}. 
To address this problem, for each beacon, we choose a set of RSSIs that are higher than or equal to 
a threshold RSSI that corresponds to $10$ meters. We replay FastSLAM for each beacon based on its set of selected RSSIs,
then get a further estimation of its location. 
Out of $11$ beacons, $8$ beacons' estimations are significantly improved, with an average $57.7\%$ improvement.  
Among the improved beacons, a set of RSSI observation locations that enclose a beacon usually gives the 
most accurate estimate. 
In our system design, we introduce a weighted entropy-based clustering algorithm to select a set of RSSI 
observation points for each beacon, shown in Section \ref{sec_clustering}.
This SLAM replay of beacon-specific high quality RSSIs, 
together with a Eulerian cycle based path planning (in Section \ref{sec_explore_planning}), 
collectively address the aforementioned design challenges.

\subsubsection{Anchor beacons improve SLAM performance}

A typical SLAM algorithm of a robot highly depends on the accuracy levels of its sensor readings 
(that measure the robot's distance/orientation from surrounding landmarks) in order to get
accurate estimations of its own location and the landmarks' locations. 
Unfortunately, the sensor readings used by our system, i.e., RSSIs of IoT devices,
are highly noisy and unreliable. To address this problem, we 
let the UAV in our system to deploy additional beacons to aid its SLAM, which are referred to 
\textit{anchor beacons}. The IoT devices to be localized are referred to as \textit{target beacons}.
We now use a simple example to illustrate the benefits of anchor beacons. 
The deployment of anchor beacons is described in Algorithm \ref{alg_search_stage} in Section \ref{sec_target_search}.
 
\begin{figure}[htb!]
	\centerline{
		\begin{minipage}{1.7in}
			\begin{center}
				\setlength{\epsfxsize}{1.8in}
				\epsffile{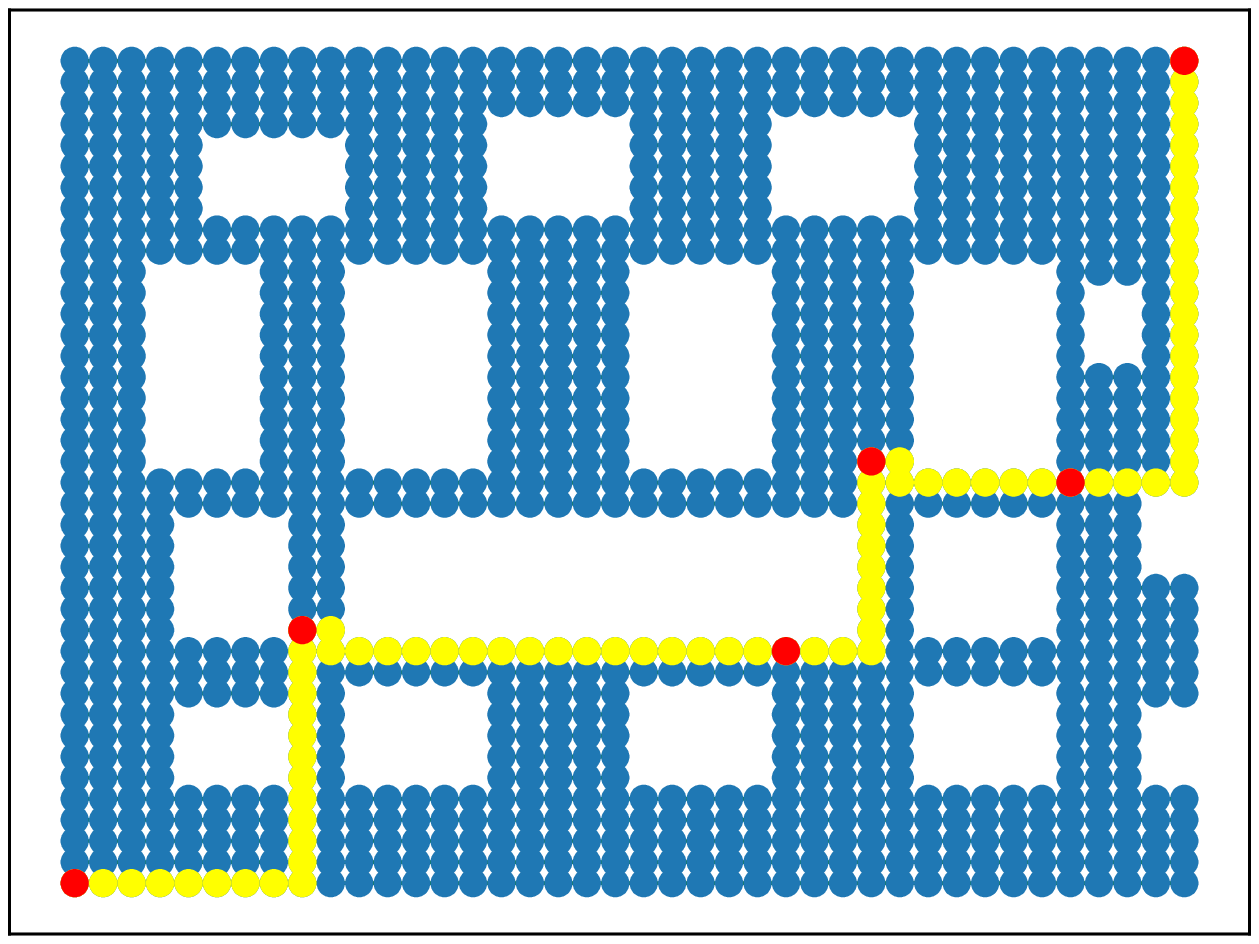}\\
				{}
			\end{center}
\caption{The locations of the five target beacons, and the UAV's path.}
\label{fig_5_targets}
		\end{minipage}
		\begin{minipage}{1.7in}
			\begin{center}
				\setlength{\epsfxsize}{1.8in}
				\epsffile{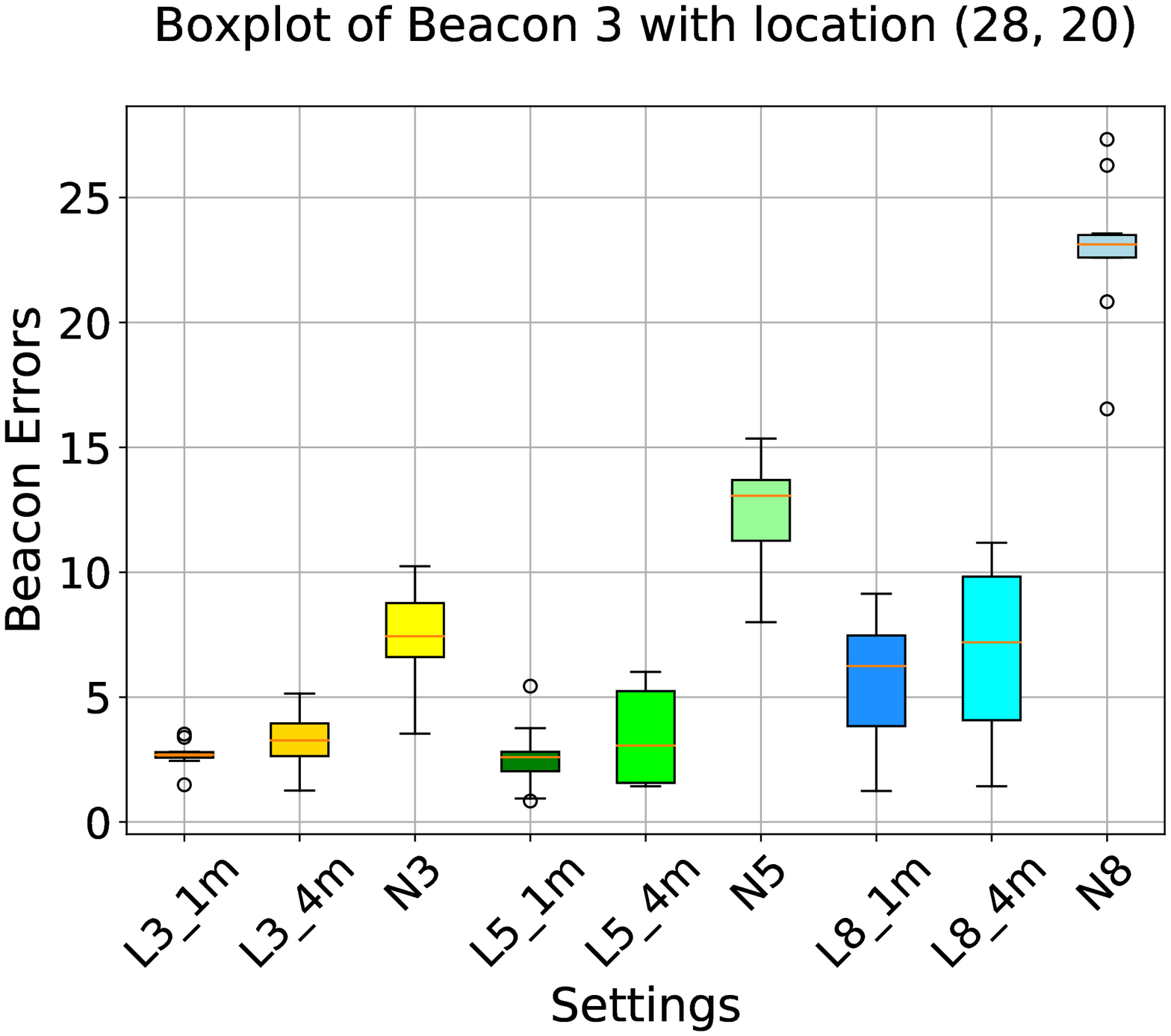}\\
			\end{center}
	\caption{Boxplots of estimation errors of the target beacon located at $(28, 20)m$.}
	\label{estimation_errors_of_3rd_target_beacons}	
		\end{minipage}
	}
	
\end{figure}

Consider an $39m$ by $39m$ area that is divided into a grid of $39$ by $39$ cells with $40$ by $40$ nodes, 
as shown in Fig. \ref{fig_5_targets}. Those $40\times 40$ nodes form a grid graph. 
The distance between any two neighboring nodes is $1m$.
We place $5$ target beacons at $5$ randomly chosen grid nodes. 
As shown in Fig. \ref{fig_5_targets}, red dots represent IoT beacons, and the yellow lines show 
the UAV's path. 
Blank blocks represent obstacles, e.g., rooms or furniture, in the area. 
The UAV uses FastSLAM while flying along the planned route. 

We simulated the environment with three different noise levels, i.e., standard deviations ($std$),
of the RSSIs generated by the target beacons:
small ($std=3$), medium ($std=5$) and large ($std=8$). 
For each noise level, we did the simulation to estimate the 5 target beacons under the following 3 cases: 
(1) Place one anchor beacon every one meter; (2) Place one anchor beacon every four meters; (3) No anchor beacon used. For each experiment, we repeated it 10 times using different random seeds. 

The simulation results show that deploying anchor beacons always gives much smaller estimation errors 
than the cases without using any anchor beacon. 
This improvement can be very significant when RSSIs are quite noisy 
or even when a target beacon is very far away from the UAV's starting point.
For example, 
Fig. \ref{estimation_errors_of_3rd_target_beacons} shows the results of the beacon at $(28,20)m$.
In the figure, L means we use anchors in the experiment, while N means we do not use them. 
Numbers 3, 5, and 8 represent the three RSSI noise levels.
The $1m$ and $4m$ respectively correspond to one anchor per one meter and one anchor per $4$ meters on the path.
We can see that compared with the case that no anchor is used, 
deploying one anchor per meter can improve the average estimation accuracy by 
$133\%$, $381\%$ and $229\%$ for RSSI noise levels being 3, 5 and 8 respectively.
In each RSSI noise level, 
we can get smaller estimation error with higher anchor density. 
When noise level is as large as 8, 
the estimation error in the case of deploying one anchor per meter is only $85\%$ of 
that when deploying one anchor every 4 meters and $22\%$ of that without using anchor beacons.

\section{System Design}\label{sec_system}

In this section, we present the design of LIDAUS. We first give an overview of the system's architecture.
Then we discuss UAV's path planing and the various components of the system.

\subsection{System Architecture}
 
\begin{figure}[htb!]
	\centerline{
		\begin{minipage}{3in}
			\begin{center}
				\setlength{\epsfxsize}{3in}
				\epsffile{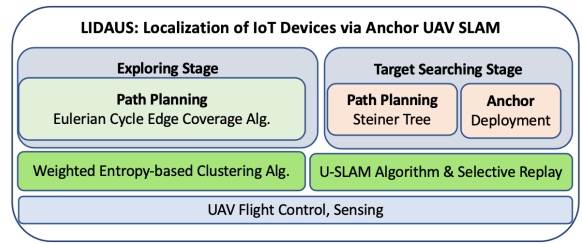}\\
				{}
			\end{center}
		\end{minipage}
	}
	\caption{System Architecture.}
	\label{fig_system}
\end{figure}

LIDAUS is a \textit{multi-stage SLAM system}
that consists of a UAV and a set of anchor beacons. The UAV is equipped with 
a wireless signal receiver module and implements a software system with its architecture shown in Fig. \ref{fig_system}.

The architecture includes the following main modules. 

\begin{enumerate}

\item \textit{Exploring stage} module.
This module generates the initial estimates of the positions of the IoT devices, 
referred to as \textit{target beacons}, in the space to be explored. 
The UAV's path planning is done through an algorithm based on a Eulerian cycle. 
The estimated 3D coordinates of target beacons are derived via
a weighted entropy-based clustering algorithm.

\item \textit{Searching stage} module. 
The UAV may need to conduct multiple searching stages in order to finalize its localization of all target beacons.
In each searching stage,  
the UAV flies along a path based on a Steiner tree that connects the estimated positions of those target beacons 
that have not been identified as \textit{found} yet. 
The UAV utilizes a dynamic anchor deployment mechanism to help localize itself during its SLAM flight. 

\item \textit{Weighted entropy-based clustering algorithm}. This algorithm runs at the end of every stage in order to 
find a cluster or set of high quality location points at which the observed RSSIs of a specific beacon 
should be used to estimate the position of the beacon. 

\item \textit{U-SLAM algorithm}. The UAV uses this 3D SLAM algorithm in its exploring and searching stages, 
and it also conducts a selective U-SLAM replay of the RSSIs found by the weighted entropy-based clustering algorithm, 
at the end of each stage.

\item \textit{UAV flight control and sensing}. The module collects RSSIs from IoT devices 
and the sensor data from the UAV's onboard height sensors. Due to space limitations, we do not describe 
the implementation details of this module in the paper.

\end{enumerate} 


\subsection{Overview of Path Planning}

We now give an overview of the UAV's path planning. 
The basic idea is to divide an indoor 3D space into multiple horizontal layers,
on which the UAV conducts SLAM trips and collects RSSI data. 
The UAV's facing direction is always aligned with the x-axis or y-axis directions of the space,
in order to minimize the movement errors caused by rotation. As for the z-axis direction, 
the quadcopter UAV used in our system always go straight up or down.
The overall workflow of LIDAUS is presented in Algorithm \ref{alg_main}. 
The UAV's trip includes an \textit{exploring stage} and one or multiple \textit{exploring stages}.

\begin{algorithm}[t]
	\caption{System Workflow}
	\hspace*{0.02in} {\bf Input:} 
	A 3D space where IoT beacons to be localized.\\
	\hspace*{0.02in} {\bf Output:} 
	Estimated locations of all IoT beacons in the space. 
	\begin{algorithmic}[1]
		\State \textit{Exploring stage} to find initial estimations of all IoT target beacons' locations. 
		The UAV invokes Algorithm \ref{alg_path_plan_exploring} for path planning. 
		It conducts U-SLAM (Section \ref{sec_u_slam}) 
		along the path and collects RSSIs. 
		At the end of the stage, it invokes weighted entropy-based clustering algorithm (in Algorithm \ref{alg_cluster})
		to find a cluster of observation locations for each target beacon, and performs
		a selective replay of U-SLAM on those clusters to derive estimated coordinates of all target beacons. 
		\State \textit{Searching stage(s)}. Multiple repeated stages may be needed in order to finalize the localizations 
		of all target beacons. In each stage, the UAV invokes Algorithm \ref{alg_search_stage} 
		for a Steiner-tree-based path planning, and then it performs U-SLAM along the path. 
		A target beacon is labeled as \textit{found} if the UAV receives a certain number of observed high RSSIs (above a threshold).
		At the end of the stage, it performs clustering and U-SLAM replay to 
		update the estimated positions of the \textit{found} target beacons.
		Conduct a new search stage if there are still not-found target beacons. 
	\end{algorithmic}\label{alg_main}
\end{algorithm}

\begin{figure}[htb!]
	\centerline{
		\begin{minipage}{1.5in}
			\begin{center}
				\setlength{\epsfxsize}{1.5in}
				\epsffile{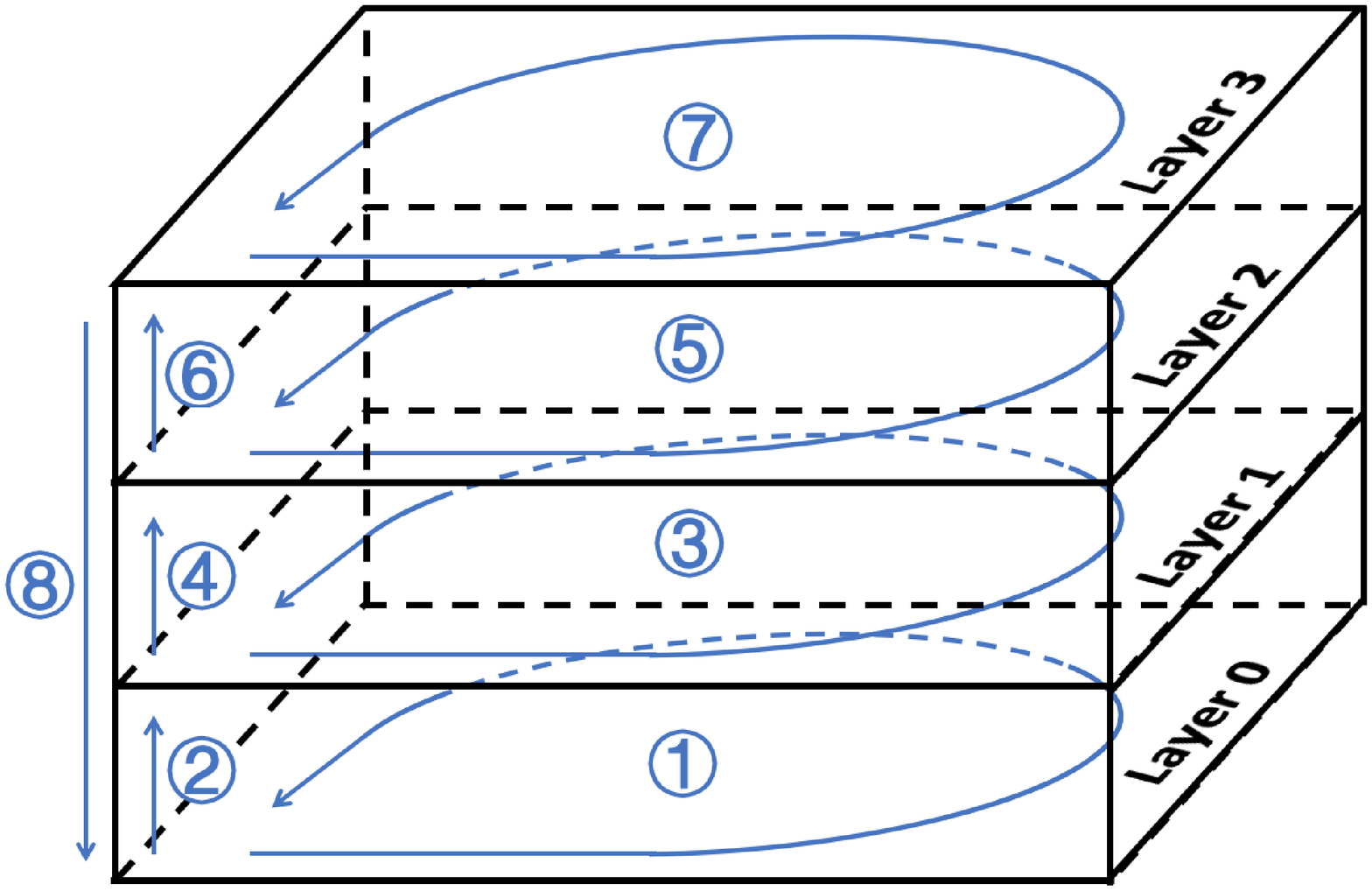}\\
				{}
			\end{center}
	\caption{Illustration of the path planning in the exploring stage.
		The UAV's path consists of $8$ segments. On each layer, its path 
		covers all edges of a grid graph representation of the layer and 
		is determined by a Eulerian cycle.}
	\label{fig_layers}
		\end{minipage}
	\hspace{2pt}
		\begin{minipage}{1.9in}
			\begin{center}
				\setlength{\epsfxsize}{1.9in}
				\epsffile{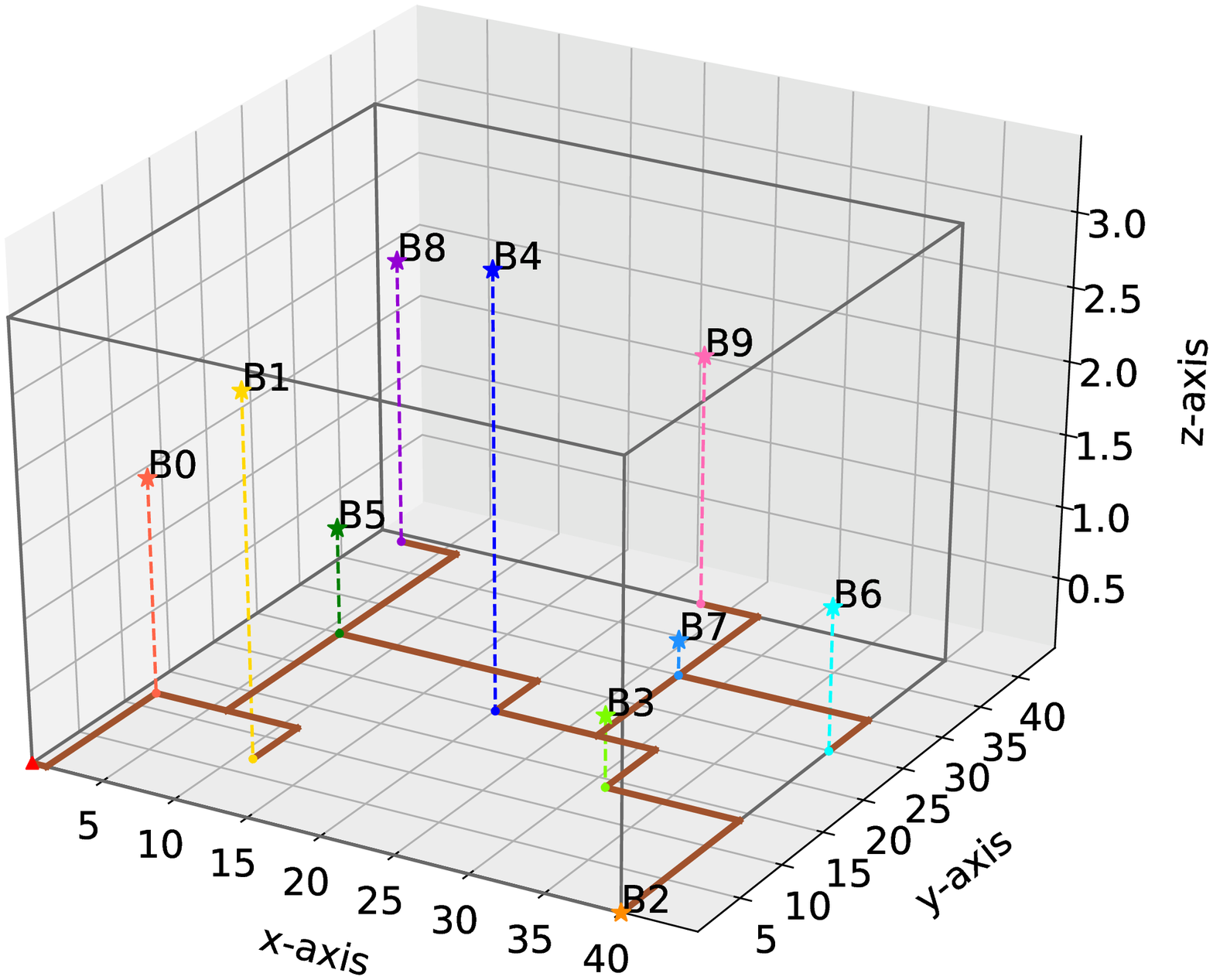}\\
				{}
			\end{center}
	\caption{An example of the path planning in a searching stage. The ten IoT beacons' estimated positions are projected on the ground layer where a Steiner tree is built to connect those projected nodes.}	
	\label{fig_tree_example}
		\end{minipage}
}

\end{figure}

In the exploring stage, the UAV flies on all the layers, one by one from a low layer to a high layer. 
On each layer, its path is determined by Algorithm \ref{alg_path_plan_exploring} (in Section \ref{sec_explore_planning}). 
Fig. \ref{fig_layers}
gives an example where the space is discretized into $4$ layers. 
The UAV starts from the origin $(0,0,0)$ and returns to the origin at the end of its trip.
Its path consists of 
$8$ segments, among which segments $1, 3, 5$ and $7$ are on the four horizontal layers. 
Since a typical indoor space has a ceiling height of $\le 3$ meters, thus this layered discretization 
along the $z$-axis will not significantly increase the complexity of our system.

In a searching stage, the estimated positions of the target beacons
whose final locations have not been decided yet are projected on the ground layer, and a Steiner tree is built to 
connect all the projected positions. Since the UAV deploys anchor beacons 
to help its SLAM process and they 
can only be deployed on the ground, thus the tree is constructed on the 
graph formed by the nodes projected to the ground layer.
The UAV keeps a record of the estimated height of each beacon, 
and when it reaches a target beacon's projected position on the ground layer, 
it flies up to reach the estimated height of the beacon. 
The algorithm to find a path 
in this stage is given in Algorithm \ref{alg_search_stage} in Section \ref{sec_target_search}. 
Fig. \ref{fig_tree_example} shows an example.

In the next three subsections, 
we will discuss U-SLAM algorithm, exploring stage, and searching stage. 

\subsection{UAV SLAM (U-SLAM)}\label{sec_u_slam}

A key component of our system is U-SLAM.
In the context of the problem we study, a UAV is a robot, 
and the wireless IoT devices to be localized are landmarks. 
U-SLAM extends FastSLAM 1.0 \cite{fastslam2003thesis} algorithm\footnote{We have found 
	that FastSLAM 2.0's performance is not as good as FastSLAM 1.0 in the problem studied 
	in this paper, as FastSLAM 2.0 relies more on the sensing data of landmarks (i.e., 
	IoT beacons' RSSIs) which are highly noisy and unreliable.}
by (1) adding a Kalman Filter (KF) to estimate the elevation of the UAV 
through a barometer and a laser sensor (ToF sensor VL53L1x)\footnote{This can be easily extended 
	by adding more accurate and longer range sensors such as mmWave sensors \cite{ti2020mmwave}.},
(2) expanding the Extended Kalman Filters (EKF) of FastSLAM to estimate the three-dimensional coordinates of
the target IoT devices, and (3) letting the UAV dynamically deploy \textit{anchor beacons} on its path 
and use the RSSIs of the anchor beacons 
in updating the posterior of its own position. 

\begin{figure}[htb!]
	\centerline{
		\begin{minipage}{1.3in}
			\begin{center}
				\setlength{\epsfxsize}{1.3in}
				\epsffile{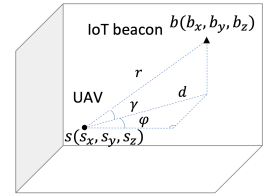}\\
				{}
			\end{center}
		\end{minipage}
	}
	\caption{An IoT beacon's relative position with respect to the UAV.}
	\label{usv_pose}
\end{figure}

Recall the IoT devices that are to be localized are called \textit{target beacons}.
Let $\mathbb{T}$ denote the set of target beacons. Let $\mathbb{A}$ denote the set of anchor beacons, 
and 
let $\mathbb{B}=\mathbb{T}\bigcup\mathbb{A}$.
Let $b_{i}$ denote the position of beacon $i$ with $b_i=(b_{i,x}, b_{i,y}, b_{i,z})$, 
and vector $\mathbf{b}$ denote the positions of all beacons. 
The UAV's position
is denoted by $s=(s_{x}, s_{y}, s_{z})$.
A beacon's relative position with respect to the UAV is given by 
$g=(r, \varphi, \gamma)^T$, i.e., the Euclidean distance, and the azimuth and elevation angle differences,
between the UAV and the beacon, as shown in Fig. \ref{usv_pose}.

\noindent\textbf{Observed and estimated parameters.}
The observed distance $r_i$ of a beacon $i$ from the UAV  
is calculated by using Eqn. (\ref{eqn_rssi}) and the observed RSSIs of beacon $i$. 
The observed height of the UAV, denoted by $h$, is from the height sensors of the UAV. 
The UAV uses a KF to estimate $s_z$ based on $h$.
Based on observed $r_i$ and estimated $s_z$ and the motion command of the UAV (denoted by $u$), 
the UAV uses a particle filter to estimate its $s_{x}, s_{y}$, and 
inside each particle, the UAV uses an EKF to estimate beacon $i$'s 
$b_i=(b_{i,x}, b_{i,y}, b_{i,z})$. Note that the RSSIs of a beacon 
do not give us observed $\gamma$ and $\varphi$.
However we can easily derive $\gamma$ after we derive the estimated $b_i$ and $s_i$.
In addition, since we keep the UAV's facing direction always along the x-axis and the y-axis of the space, 
the azimuth angle difference $\varphi$ can also be derived from estimated $b_i$ and $s$.

Let $r_{i,t}$ denote the observed distance between beacon $i$ and the UAV at time $t$.
Vector $\mathbf{r}_{t}=(r_{1,t}, r_{2,t}, ..., r_{i,t},..., r_{|\mathbb{B}|,t} )$ 
includes all beacons at time $t$.
The UAV's location
at time $t$ is denoted by $s_t$,
and an IoT beacon $i$'s position at time is denoted by $b_{i,t}$.
Let $u_t$ denote the motion command given by the UAV's controller to the UAV
at time $t$.
The UAV's history of positions from time $0$ to time $t$ is denoted by $s_{0:t}$. 
Similarly, we have $\mathbf{r}_{0:t}$, $h_{0:t}$, and $u_{0:t}$. 

\subsubsection{\textbf{Calculate the posteriors of the positions of UAV and beacons}}
The UAV updates its estimation of the following probability 
during its flight\footnote{We do not need to solve the data association problem as in 
traditional robotics SLAM algorithms, as each beacon's signal contains its unique ID.}.
\begin{equation}\label{eqn1}
p(s_{0:t}, \mathbf{b})|\mathbf{r}_{0:t}, h_{0:t}, u_{0:t})
\end{equation}
At the end of the UAV's flight, the final estimate of the mean of $\mathbf{b}$ tells us
the estimated locations of all beacons. 

The UAV solves (\ref{eqn1}) via the conditional 
independence between beacons based on the knowledge of the UAV's path. That is,  
\begin{equation}\label{eq2}
\left.\begin{aligned}
&p(s_{0:t}, \mathbf{b} | \mathbf{r}_{0:t},h_{0:t}, u_{0:t})\\
&=p(s_{0:t}| \mathbf{r}_{0:t},h_{0:t}, u_{0:t}) p(\mathbf{b} | s_{0:t}, \mathbf{r}_{0:t},h_{0:t}, u_{0:t})\\
&=p(s_{0:t}| \mathbf{r}_{0:t},h_{0:t}, u_{0:t})
\prod_{i=1}^{|\mathbb{B}|} p(b_i | s_{0:t}, r_{i,0:t}, h_{0:t}, u_{0:t})
\end{aligned}\right.
\end{equation}

The algorithm estimates the UAV's $s_x$ and $s_y$ 
by using a particle filter that has $L$ particles, 
and it estimates $s_z$ by using a KF with observed height $h$ from height sensors. In addition, the algorithm
estimates beacon $i$'s posterior $ p( b_i| s_{0:t}, r_{i,0:t}, h_{0:t}, u_{0:t})$ 
by using EKFs, with one EKF for each beacon. 

Consider a beacon $i$.
The algorithm uses the following formula to compute the posterior  
of its position $b_i$:
\begin{eqnarray}
&p(b_i | s_{0:t}, r_{i,0:t}, h_{0:t}, u_{0:t}) \nonumber \\
&=\eta \cdot p( r_{i,t} | b_i, s_t )\cdot p( b_i | s_{0:t-1}, r_{i,0:t-1}, h_{0:t}, u_{0:t-1})  
\end{eqnarray}\label{eqn_ekf}
where $\eta$ is a normalizing factor.

The algorithm maintains a set of particles $\mathbf{Y}_t$
and a KF at time $t$. 
Each particle $\ell$ contains its estimated x- and y-coordinates of the UAV at time $t$ 
(i.e., $s^{[\ell]}_x, s^{[\ell]}_y$).
A KF contains estimated UAV's height $s_z$. 
Let $s^{[\ell]}_t$ denotes the vector collection of these three estimates.
In addition, particle $\ell$ contains the estimated mean and covariance of the position of each beacon $i$ at time $t$,
denoted by $(\mu_{i,t}^{[\ell]},\Sigma_{i,t}^{[\ell]})$.
For ease of exposition, we drop notations $i, t, \ell$ 
and consider a beacon's position distribution with mean and covariance $(\mu,\Sigma)$, then 
we have $\mu = (\mu_x, \mu_y, \mu_z)$
and 
\begin{equation*}
\Sigma = 
\begin{pmatrix}
\sigma^2_x & \sigma_{x,y} & \sigma_{x,z}  \\
\sigma_{x,y} & \sigma^2_{y} & \sigma_{y,z}  \\
\sigma_{x,z} & \sigma_{y,z} & \sigma^2_{z} 
\end{pmatrix}
\end{equation*}

Let $\mathbf{Y}_{t}^{[\ell]}$ denote particle $\ell$ at time $t$, 
then $\mathbf{Y}_{t}^{[\ell]}=\{s_{t}^{[\ell]},(\mu_{1,t}^{[\ell]},\Sigma_{1,t}^{[\ell]}),...,
(\mu_{|\mathbb{B}|,t}^{[\ell]},\Sigma_{|\mathbb{B}|,t}^{[\ell]})\}$, and let 
$\mathbf{Y}_t=\{ \mathbf{Y}_{t}^{[1]}, ..., \mathbf{Y}_{t}^{[L]} \}$.

\subsubsection{\textbf{The iterative steps}}

The algorithm generates  $\mathbf{Y}_t$ from $\mathbf{Y}_{t-1}$ at time $t$ 
based on the latest motion control $u_t$ and observed values $\mathbf{r}_t$ and $h_t$ by running the following steps: 
\begin{enumerate}
	\item \textbf{Step 1}. The height KF estimates the UAV's $s_{z,t}$ at $t$. 
	\item \textbf{Step 2}. Each particle $\ell$ predicts the new position $s^{[\ell]}_t$.  
	based on $u_t$, $\mathbf{s}^{[\ell]}_{t-1}$, and $s_{z,t}$
	\item \textbf{Step 3}. Update the EKF estimate of each beacon in every particle.
	\item \textbf{Step 4}. Calculate the importance weight of each article.
	\item \textbf{Step 5}. Conduct an importance re-sampling to derive a new  particle set. 	
\end{enumerate}

\subsubsection{\textbf{Impacts of RSSI noise and UAV's motion noise}}

There are two types of measurements that can impact the estimation accuracy of the beacons' positions
and the UAV's position over time. They are the beacons' RSSI noise and the UAV's motion noise.
We next examine all 5 steps carefully. 

Step 1's height estimation is affected by the height sensors' noises and the UAV's motion command noise.  

Step 2 of the algorithm utilizes a probabilistic motion model to predict the UAV's position, i.e., 
$s^{[\ell]}_t \sim p(s_t | u_t, s^{[\ell]}_{t-1})$, so this prediction is affected by the 
motion control noise, not the noise of the measured RSSIs of beacons.

Step 3 updates the position estimation of a beacon $i$ by using an EKF. Consider 
particle $\ell$, its EKF first calculates a predicted distance $\hat{r}_i$ from
beacon $i$ with the beacons's position estimated  
at the previous time step $t-1$ and the predicted new position of the UAV, i.e., 
\begin{equation}
\hat{r}^{[\ell]}_{i,t} = g(s_t^{[\ell]}, \mu^{[\ell]}_{i, t-1})
\end{equation}
where $g(s, b_i)$ is the Euclidean distance function between $s$ and $b_i$.

Then the EKF derives the estimated mean and covariance of beacon $i$ at time $t$ according to the following
equations:
\begin{eqnarray}
\mu^{[\ell]}_{i, t} &=& \mu^{[\ell]}_{i, t-1} + K^{[\ell]}_t(r_{i,t}-\hat{r}^{[\ell]}_{i, t-1}) \\
\Sigma_{i,t}^{[\ell]} &=& (I - K^{[\ell]}_{i,t} G^{[\ell]}_{i,t}) \Sigma_{i,t-1}^{[\ell]} 
\end{eqnarray}
where coefficients $K^{[\ell]}_{i,t}$ and $G^{[\ell]}_{i,t}$ both are functions of 
$s_t^{[\ell]}, \mu^{[\ell]}_{i, t-1}, \Sigma_{i,t-1}^{[\ell]} $.
Therefore we can write the following functions to calculate the estimated mean and covariance of
beacon $i$ in particle $\ell$ at time $t$:
\begin{eqnarray}
\mu^{[\ell]}_{i, t} &=& f(r_{i,t},s_t^{[\ell]}, \mu^{[\ell]}_{i, t-1}, \Sigma_{i,t-1}^{[\ell]} ) 
\label{eqn_mu_rec} \\
\Sigma_{i,t}^{[\ell]} &=& f(s_t^{[\ell]}, \mu^{[\ell]}_{i, t-1}, \Sigma_{i,t-1}^{[\ell]} ) 
\label{eqn_sigma_rec}
\end{eqnarray}

Step 4 calculates the importance weight of each particle based on the measured RSSIs of the beacons at 
time $t$. Specifically the weight of particle $\ell$ is a function of $r_{i,t}$.
\begin{equation}
w^{\ell}_t = f(r_{i,t},s_t^{[\ell]}, \mu^{[\ell]}_{i, t-1}, \Sigma_{i,t-1}^{[\ell]} ) \label{eqn_weight}
\end{equation}

We follow the standard procedure to design the KF, the EKFs, and the importance sampling for particle filters.
The design details are not included here due to space limitations.

\noindent \textbf{Remarks.}
Equations (\ref{eqn_mu_rec}), (\ref{eqn_sigma_rec}), and  (\ref{eqn_weight}) show that the accuracy of estimated 
	positions of the UAV and of beacons are highly dependent on the accuracy of the measured RSSIs.
	Thus, we let the UAV deploy anchor beacons on its path and only use anchor beacons' RSSIs in the estimation 
	of its own locations in its U-SLAM, which helps improve its 
	U-SLAM's localization accuracy.

\subsection{SLAM Selective Replay}\label{sec_slam_replay}
At the end of each stage (either the exploring stage or each searching stage), 
for a beacon $i$, the UAV selects a set of RSSI points according to a weighted entropy-based clustering method 
(i.e., Algorithm \ref{alg_cluster} in Section \ref{sec_clustering}), denoted by $C_i$. 
Then the UAV conducts an offline replay of U-SLAM  based on $C_i$ to update its estimation of beacon $i$'s position. 

\subsection{Weighted entropy-based clustering algorithm}\label{sec_clustering}

We design this algorithm to select a set or cluster of position points (denoted by $C_i$) for a beacon $i$,
at which the observed RSSIs of beacon $i$
can be used for a SLAM selective replay to get an accurate estimate of beacon $i$'s position. 
This algorithm is shown in Algorithm \ref{alg_cluster}. 
In addition, the center of cluster $C_i$ is also used as the initial estimate in the SLAM selective replay for beacon 
$i$ at the end of the exploring stage.

Let $P_{init}=\{p_1, ..., p_N \}$ denote the set of the observation points or positions 
where RSSIs are collected in a trip of the UAV.
For an observation point $p_j$, the UAV derives a set of the medians of observed RSSIs, 
denoted by $R^j=\{R^j_1, ..., R^j_M \}$ for 
all $M$ IoT beacons, denoted by set $T=\{T_1, ..., T_M \}$.

\begin{algorithm}[t]
	\caption{Weighted Entropy-based Clustering Algorithm}
	{\bf Input:} 
	(1) $P_{ini}=\{p_1, ..., p_N \}$;
	(2) $R^j=\{R^j_1, ..., R^j_M \}$, $M$ IoT beacons $T=\{T_1, ..., T_M \}$;
	(3) Initial RSSI difference level $R_{dL}=0$;
	(4) Initial RSSI cluster threshold $R_{th,c}$;
	(5) Entropy threshold $S_{th}$;
	(6) Edge weight look-up table of each 3D cube $\{G_1, ..., G_q \}$
	(7) $L_G$.
	(8) 
	$L_{i, k}$.
	Let $E_{i,k}$ denote the set of points on edge $e_k$ that are selected for beacon $i$. \\
	{\bf Output:} 
	$\forall i$, cluster $C_i$ and its center $\hat{b}_i$.
	\begin{algorithmic}[1]
		\State INITIALIZE:		
		Entropy of target beacons $\{S_1=0, ..., S_N=0 \}$, $C_i=\emptyset$, $L_{i, k}=0$, $E_{i,k}=\emptyset$.	
		\ForAll{beacon $i$ in $T$} 
		\State $S_{i,prev}=0, S_{i}=1, \forall i$
		\While{$S_i < S_{th}$ and $C_i \subset P_{init}$}
		\If{$S_i \le S_{i, prev}$ } 
		\State Decrements $R_{th,c}$ by $1$.
		\Else if not the first iteration
		\State Increments $R_{dL}$ by $1$.
		\EndIf
		\State Reset entropy $S_i = 0$
		
		\ForAll{$p_j$ in $P_{ini}$} 
		\State $R^j_{max} = \underset{i}{\max} \{R^j_1, ..., R^j_i,..., R^j_N\}$;			
		\If{$R^j_i == R^j_{max}$ or $|R^j_i - R^j_{max}|\leq R_{dL}$  }
		\If{$R^j_i \ge R_{th,c}$ }
		\State $C_i= C_i \bigcup \{p_j\}$
		\EndIf
		\EndIf
		\EndFor
		\State Calculates the 1-means center position $c_i$ for $C_i$ 
		\State \hspace{6pt} and locates the cube $G_q$ where $c_i$ is in.
		\ForAll{$p_j$ in $C_i$}
		\State Find the edge $e_{k}$ where $p_j$ is located and 
		\State \hspace{6pt}  $E_{i,k}= E_{i,k} \bigcup \{p_j\}$, $L_{i, k}=L_{i, k}+1$.
		\EndFor
		\State $L_i=\Sigma_k L_{i,k}$
		\ForAll{$E_{i,k}$ of $C_i$ }
		\State Find the weight $w_{k}$ of $E_{i,k}$ 
		\State \hspace{6pt} from the weight look-up table of $G_q$.
		\State Update entropy: 
		\State \hspace{6pt} $S_i = S_i + (- w_k \frac{L_{i,k}}{L_i} \log( \frac{L_{i,k}}{L_i}) )/\log(L_G)$
		\EndFor
		\EndWhile
		\State $\hat{b}_i = c_i$
		\EndFor
		\State \Return $C= \{C_1, ..., C_M \}$ and $ \{\hat{b}_1, \hat{b}_2, ..., \hat{b}_M\}$.
	\end{algorithmic}\label{alg_cluster}
\end{algorithm}

For a beacon $i$, 
we use two thresholds to control the selection of RSSI observation points to be added into cluster $C_i$: 
RSSI difference level $R_{dL}$ 
and RSSI cluster threshold $R_{th,c}$. These two thresholds are dynamically adjusted during the execution 
of the clustering algorithm in order to get a sufficient number of points in $C_i$ 
and in the mean time to ensure only location points 
with high quality RSSIs for the beacon are added into $C_i$.

Recall that based on the relationship between RSSI and distance, 
in general the closer a location is to a beacon, the stronger its RSSI can be observed. 
Thus we decide that the points where the RSSIs observed for a beacon is no less than $R_{th,c}$ 
can be candidate points for the beacon. 
However, due to multi-path and other factors, 
a strong RSSI of a beacon at a location that is not so close to the beacon might also be observed. 
To mitigate this impact, we assign a location point as a high quality observation location 
only to the beacon that has the strongest RSSI among 
all beacons with observed RSSIs at the location, i.e., the location is only used for 
the beacon with strongest RSSI, not used for other beacons' estimation 
even though their RSSIs are also observed.  
In case we cannot collect sufficient observations for a beacon, we lessen this strongest assignment requirement by assigning a location to a beacon if its RSSI at the location is within a difference threshold $R_{dL}$ 
from the strongest RSSI
at the location. This is shown on line $12$ of Algorithm \ref{alg_cluster}.

\begin{figure}[htb!]
	\centerline{
	\begin{minipage}{1.8in}
			\begin{center}
				\setlength{\epsfxsize}{1.8in}
				\epsffile{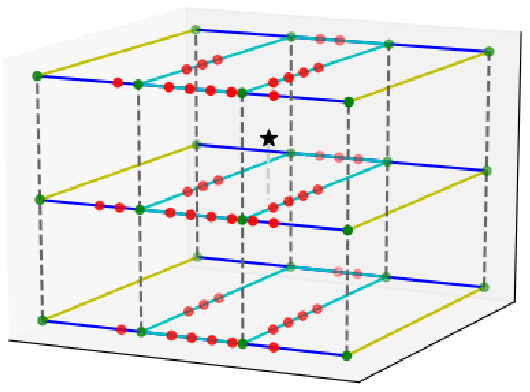}\\
				{}
			\end{center}
	\caption{A toy example to calculate weighted entropy of the cluster of points (the red dots in the figure) of a beacon 
		(the black dot in the figure).}
	\label{fig_entropy_example}
	\end{minipage}	
	\begin{minipage}{1.3in}
			\begin{center}
				\setlength{\epsfxsize}{1.3in}
				\epsffile{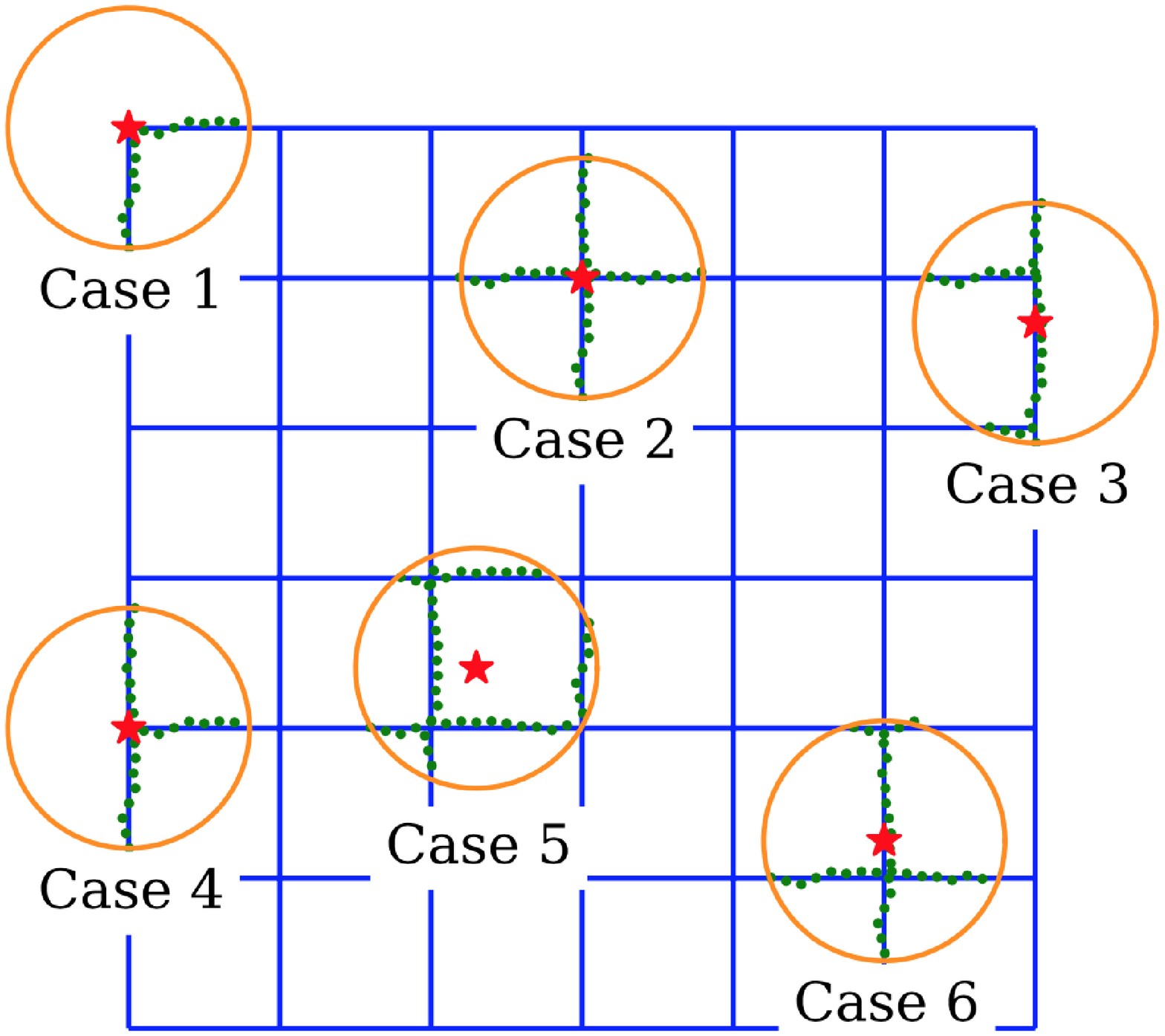}\\					
				{}
			\end{center}
	\caption{Example locations when a beacon is on a 2D grid graph. Red stars represent beacons.
		The points inside a circle and along edges 
		are the locations that can be used by SLAM replay.}
	\label{grid_beacon_cases}
	\end{minipage}
	}
\end{figure}

In addition, 
we use Shannon entropy to evaluate the quality of the selected observation points for a beacon. 
Recall all observation points are along the edges
of a grid graph. For a beacon $i$, recall $C_i$ denotes the set of the selected points for this beacon. 
Those points in $C_i$ are distributed on various edges. 
We use the following two methods to evaluate the quality of those points in terms of estimating beacon $i$'s
position. 
\textbf{First}, note that the more evenly distributed the points on those edges, the better 
location estimation we can get for this beacon. For example, consider case 5 in Fig. \ref{grid_beacon_cases}.
If all observation points are evenly distributed on all four edges instead of on only one edge, 
our SLAM algorithm can get high estimation accuracy.  
We use an entropy metric to measure the level of evenness or balance of the points assigned to beacon $i$.
\textbf{Second}, for a particular beacon, 
we associate a weight $w_k$ to each edge $e_k$; the $w_k$ of an edge gets smaller if the edge 
is further away from the beacon. Specifically, we propose to use the following 
weighted entropy of beacon $i$'s RSSI observation points to evaluate the quality of cluster $C_i$:
\begin{equation}
S_i = \Sigma_{k} (- w_k \frac{L_{i,k}}{L_i} \log( \frac{L_{i,k}}{L_i}) )/\log(L_G)
\end{equation}
where $L_{i, k}$ denotes the number of observation points in an edge $e_k$ that is selected for beacon $i$,
$L_i=\Sigma_k L_{i,k}$, and $L_G$ denotes the total number of edges of the multi-layer grid graph representation 
of the search space.
The goal of Algorithm \ref{alg_cluster} is to find a $C_i$ for a beacon $i$ so that the $C_i$'s weighted entropy $S_i$ is above 
a quality threshold $S_{th}$.
Note that since the edges are weighted, we need to multiply the above $S_i$ with another scaling factor to scale 
an entropy in the range of $[0,1]$.  
Note that a simple clustering to find location estimates of iBeacons is used in \cite{le2018sombe}, but their method is 
simply a direct application of K-means clustering. 

Fig. \ref{fig_entropy_example} shows a toy example to illustrate the calculation of the weighted entropy of the cluster of positions
(shown as the red dots in the figure) for a beacon (shown as the black dot in the figure). 
There are $48$ observation points in this beacon’s cluster. 
There are $30$ edges in this space.
Given a target estimated located in the middle cube of the graph, 
$12$ cyan edges are the nearest (or core) edges to the beacon, with weight $w_C$; 
$12$ blue edges are adjacent to the core edges, and their weights are $w_A$; 
$6$ yellow edges belong to the Other class with weight $w_O$. 
The $12$ core edges have $40$ observation locations in total, 
the 12 adjacent edges have 8 observations, and there are 0 observation locations in other edges. 

\subsection{Exploring Stage Path Planning}\label{sec_explore_planning}

In this stage, the UAV models each horizontal layer as a grid graph. 
Then it travels along a path based on an Eulerian cycle to cover all edges of the grid graph
on each horizontal layer. It visits all layers, one by one from lower layers to higher layers, as shown in 
Fig. \ref{fig_layers}. 
It runs U-SLAM during its trip and in the mean time it collects RSSIs of all IoT devices while flying. 
After returning back, the UAV uses a \textit{weighted entropy-based clustering algorithm} to derive 
initial estimated positions of the IoT devices in the space.

\noindent {\textbf{Path planning based on Eulerian cycle}}.
On each layer, the UAV follows a path determined by Algorithm \ref{alg_path_plan_exploring}, 
which is a minimum distance path that 
\textit{covers all the edges} of the squares of the layer.
Let $G_{in}$ denote the grid graph on a layer. 
Assume that each edge of the graph has a unit cost.  
The basic idea of Algorithm \ref{alg_path_plan_exploring} is to first convert $G_{in}$ to an even-degree graph
(i.e., every node has even degree), and then utilize an Eulerian cycle algorithm to find the 
least cost (i.e., shortest distance) tour that covers all edges of the graph. The tour starts from the origin
of $G_{in}$ and returns back to the origin at the end of the tour. 

\noindent \textbf{Rationale for collecting RSSIs on all edges of the grid graph representation of a horizontal layer.}
This design addresses the problem of direction ambiguity of a beacon
due to the fact that RSSIs can only indicate distance, but no direction or angle information.
For ease of exposition, we take a look at a 2D plane shown Fig. \ref{grid_beacon_cases},
where the red dots show the possible locations of an IoT beacon.
We see that in all cases, an edge covering tour can collect RSSI data of a beacon along both x-axis and y-axis directions.
In addition, for case 5 (a very common case),
all the positions where we get the beacon's RSSI data enclose the beacon. 
This design can greatly improve the localization accuracy of U-SLAM. 
\begin{algorithm}[t]
	\caption{Path Planning of Exploring Stage}
	\hspace*{0.02in} {\bf Input:} 
	Grid graph $G_{in}$ on a plane\\
	\hspace*{0.02in} {\bf Output:} 
	UAV's initial fly path $P_0$
	\begin{algorithmic}[1]
		\If{$G_{in}$ is not an even-degree graph}
		\State Form a graph $G^*$ from all odd-degree nodes in $G_{in}$
		\ForAll{node pairs $u, v$ in $G^*$}
		\If{$e_{uv} \notin G_{in}$}
		\State Add $e_{uv}$ in $G^*$ 
		\State Set weight $w(e_{uv})=d_{min}(u,v)$, i.e., the
		\State  \hspace{4pt}  shortest path distance between $u$, $v$ in $G_{in}$.
		\EndIf
		\EndFor
		\State Use Blossom V algorithm \cite{kolmogorov2009blossom} to find the minimum 
		\State \hspace{4pt} cost perfect matching $M^*$ in $G^*$.
		\State Add edges in $M^*$ into $G_{in}$ to get a new graph $G_{in}^*$, 
		\State \hspace{4pt} which is an even-degree graph.
		\EndIf
		\State Use an Eulerian cycle algorithm (e.g., Fleury's algorithm)  to find a least-cost tour $P_0$
		in $G_{in}^*$.\\
		\Return $P_0$
	\end{algorithmic}\label{alg_path_plan_exploring}
\end{algorithm}

\subsection{Searching Stage}\label{sec_target_search}

Following the exploring stage, the UAV goes through a limited number of target searching stages
$Stage_1, Stage_2, ..., Stage_K$. The limit $K$ is a user tunable parameter, depending on the specific application scenario.
This stage is described in Algorithm \ref{alg_search_stage}.

In each stage, the UAV constructs a Steiner tree \cite{hwang1992steiner} to reach all remaining 
not-yet-found target devices, based on the latest estimated positions of them from previous stages. 
It travels along the shortest branch of the tree to fly close to the targets on that branch.
While flying
along the tree nodes, the UAV also deploys anchor beacons to help its U-SLAM computation. 
During this stage,
the UAV updates its estimation of all targets through U-SLAM.  
The rationale of utilizing anchor beacons is to improve the estimation accuracy of U-SLAM,  
as the algorithm (and in fact all SLAM algorithms based on our knowledge) 
is highly dependent on the measured signals of beacons, i.e., RSSIs,
but RSSIs are not reliable.

In addition, since a Steiner tree is a minimum-weight tree that spans all target device nodes in the graph of the area, the Steiner tree path can minimize the the deployment cost of anchor beacons.
Note that we let the UAV fly close to a target, as RSSIs are more reliable when a receiver is very close to a transmitter. 
A target is labeled as \textit{found} when the UAV observes a sufficient number of very strong RSSIs. 

After exploring the shortest branch of the Steiner tree in \textit{Stage $i$}, for all targets that are labeled as \textit{found},
the UAV runs a U-SLAM selective replay based on the set of RSSI observation locations selected by the entropy clustering algorithm.  
If there are still targets that are not labeled as found yet, the UAV starts a new stage again. 
The total number of stages is bounded by a empirically determined number\footnote{Note that in our simulations and experiments, the UAV can always quickly label all targets as found within 5 stages.}.

\begin{algorithm}[t]
	\caption{Searching Stage with Anchor Steiner Tree}
	\hspace*{0.02in} {\bf Input:} 
	(1) Initial estimated positions of target beacons' coordinates $\mathbf{\hat{b}}^0 = \{\hat{b}_1^0, ...,\hat{b}_M^0 \}$,
	(2) $R_{th}$, threshold RSSI value,
	(3) $n_{th}$, threshold, i.e., a target beacon is labeled as \textit{found} if the number of its RSSIs that is above $R_{th}$
	reaches  $n_{th}$. \\
	\hspace*{0.02in} {\bf Output:} 
	Target beacons' estimated $\mathbf{\hat{b}} = \{\hat{b}_1, ...,\hat{b}_M \}$.
	\begin{algorithmic}[1]
		\State INITIALIZE:		
		UAV starts at origin: $(0, 0, 0)$, let $\mathbf{\hat{b}} = \mathbf{\hat{b}}^0 $,
		found target set $T_{found} = \emptyset$, not-yet-found target set $T_{non} = \mathbb{T}$, 
		set of the coordinates of expected anchors $A_{exp} = \emptyset$, set of the coordinates of deployed anchors $A_{dep} = \emptyset$, 
		set of tree nodes $N_{tree} = \emptyset$.

		\State Discretize the ground plane as a grid graph $G$ with each cell's $SideLength=1m$.
		\While{$T_{non} \ne \emptyset$} 	
			\State Project every target beacon $t_i$'s estimated position $(\hat{b}_{i,x}, \hat{b}_{i,y}, \hat{b}_{i,z})$ 
					  in $T_{non}$ to its closest node on $G$, denoted by $(v_{i,x}, v_{i,y})$, 
					  and put all these nodes in $T_{non}^{proj}$
			\State $N_{tree} = T_{non}^{proj} + A_{dep}$
			\State Use UAV's current position as root to build a Steiner tree $Tree$ out of $G$ that connects all nodes in $N_{tree} $.
			\State Let path $P =$ the shortest branch of $Tree$.
			\State Generate $A_{exp}$, the set of expected positions of the anchors to be deployed on $P$.
			
			\State $F = \emptyset$	
			\State UAV flies on the ground plane, following path $P$ and performing U-SLAM algorithm.		
			\ForAll{step point $p_j \in P$}
				\If{$p_j \in A_{exp}$ and $p_j \notin A_{dep}$}
					\State UAV deploys one anchor beacon on the ground at $p_j$
					\State $A_{dep} = A_{dep} + \{p_j\}$
				\EndIf
				\If{$p_j$ and a node $v_i \in T_{non}^{proj}$ have the same x and y coordinates}
					\State UAV flies up to $(\hat{b}_{i,x}, \hat{b}_{i,y}, \hat{b}_{i,z},)$ of $\hat{b}_i \in T_{non}$, 
							  as $v_i$ is $\hat{b}_i$'s projection, then flies around, and finally down to $p_j$
				\EndIf
				\If{there are $n_{th}$ collected RSSIs $R_i^j$ for target beacon $t_i$ such that $R_i^j >R_{th}$} 
					\State $F = F  + \{t_i\}$
				\EndIf								
			\EndFor
			\State $T_{found} = T_{found}  + F$
			
			\State Use Algorithm \ref{alg_cluster} based on all RSSIs collected so far to get cluster $C_i$ for each target beacon $t_i$.
			\ForAll{$t_i \in T_{non}$ }
				\State Use $C_i$ and $R^j_i$ to replay U-SLAM algorithm
				\State Update $\hat{b}_i$ in $\mathbf{\hat{b}}$
			\EndFor
			\State $T_{non} =T_{non}  - F$
						
		\EndWhile
				
		\State 
		\Return $\mathbf{\hat{b}}$
	\end{algorithmic}\label{alg_search_stage}
\end{algorithm}


\section{Performance Evaluation}\label{sec_eval}

In this section, we show the effectiveness of LIDAUS through simulations and experiments.
Due to space limitations, we only report some results here. 

\subsection{Simulation settings}

We simulate the operation of LIDAUS in a 3D space  with dimensions $40m \times 40m \times 3m$. 
We discretize it into $4$ horizontal layers, and their heights are of $0m$, $1m$, $2m$ and $3m$ respectively. 
We model each layer as a grid graph with squares of size $10m \times 10m$. 
There are $10$ IoT devices (i.e., target beacons) scattered in the space as shown in Fig. \ref{fig_tree_example}. 
The UAV has a Gaussian motion noise with standard deviation ($std$) of $0.05m$ in horizontal direction and $std=0.03m$ in vertical direction, 
and the RSSI data received from a target beacon is generated through Eqn. (\ref{eqn_rssi}) with a Gaussian noise $std=0.1$.
In simulations, the UAV always starts at the origin $(0, 0, 0)$, and it collects RSSIs at every $1m$ step.

We compare LIDAUS with two baseline methods: 
\textit{Random SLAM Fly} and \textit{Naive SLAM Search}.
With these two baseline methods, the UAV also starts at the origin and measures RSSIs every $1m$, and it also 
utilizes the 3D U-SLAM algorithm. But these two baseline methods 
do not deploy any anchor beacon. 
For all three methods, the RSSI threshold to label a target beacon as \textit{found} is $-51$, which corresponds to  
a distance of $1.7m$ (i.e., about the length of the diagonal line of a cube with $1m$ side length\footnote{The relationship between RSSI and distance varies across different types of devices. 
	A future work of ours is to design a learning module to 
	deal with a mixture of various different types of wireless signals.}). 
The UAV in LIDAUS follows the design given in Section \ref{sec_system}.

With \textit{Random SLAM Fly}, the UAV randomly chooses a direction to move at each step of $1m$.
The UAV will stop if all targets have been labeled as found, or it will stop if 
its total flying step reaches $10k$ steps.
Note that this Random SLAM Fly is similar to HumanSLAM \cite{bulten2016human},
	but the former works in a 3D space and the motion control estimation is more accurate than the smartphone-based
	human motion estimation in HumanSLAM.
	
With \textit{Naive SLAM Search}, the UAV also conducts a multi-stage U-SLAM flight. In its exploring stage, 
it flies layer by layer in the same discretized space as LIDAUS, but it follows the 8-like shape path (same as the 
one shown in Fig. \ref{fig_targets_multiple}) on each layer in order to resolve direction or angle ambiguity. 
After that, the UAV sets the estimated position of the closest unfound target as its destination and flies toward it. 
During this trip, it keeps updating the estimates of all unfound target beacons via U-SLAM as new RSSIs are being observed
constantly, 
and when the UAV finds the received RSSI from a target is greater than $-51$ at least three times, 
it will consider that target as \textit{found}. 
When UAV reaches the destination, it will find the closest estimated position of an unfound target as the next destination 
and repeat the above process until all target beacons are labeled as found, or the total flying steps reach $10k$ steps. 
To avoid keeping searching the same target all the time or getting into an infinite loop, we set a counter 
for each target. Whenever a target's estimated position is chosen as a destination, its counter increments by one. 
If a target's counter reaches $5$, it will not be chosen as a destination in future. 

\subsection{Comparison between methods}

\begin{figure}[htb!]
	\centerline{
		\begin{minipage}{1.7in}
			\begin{center}
				\setlength{\epsfxsize}{1.7in}
				\epsffile{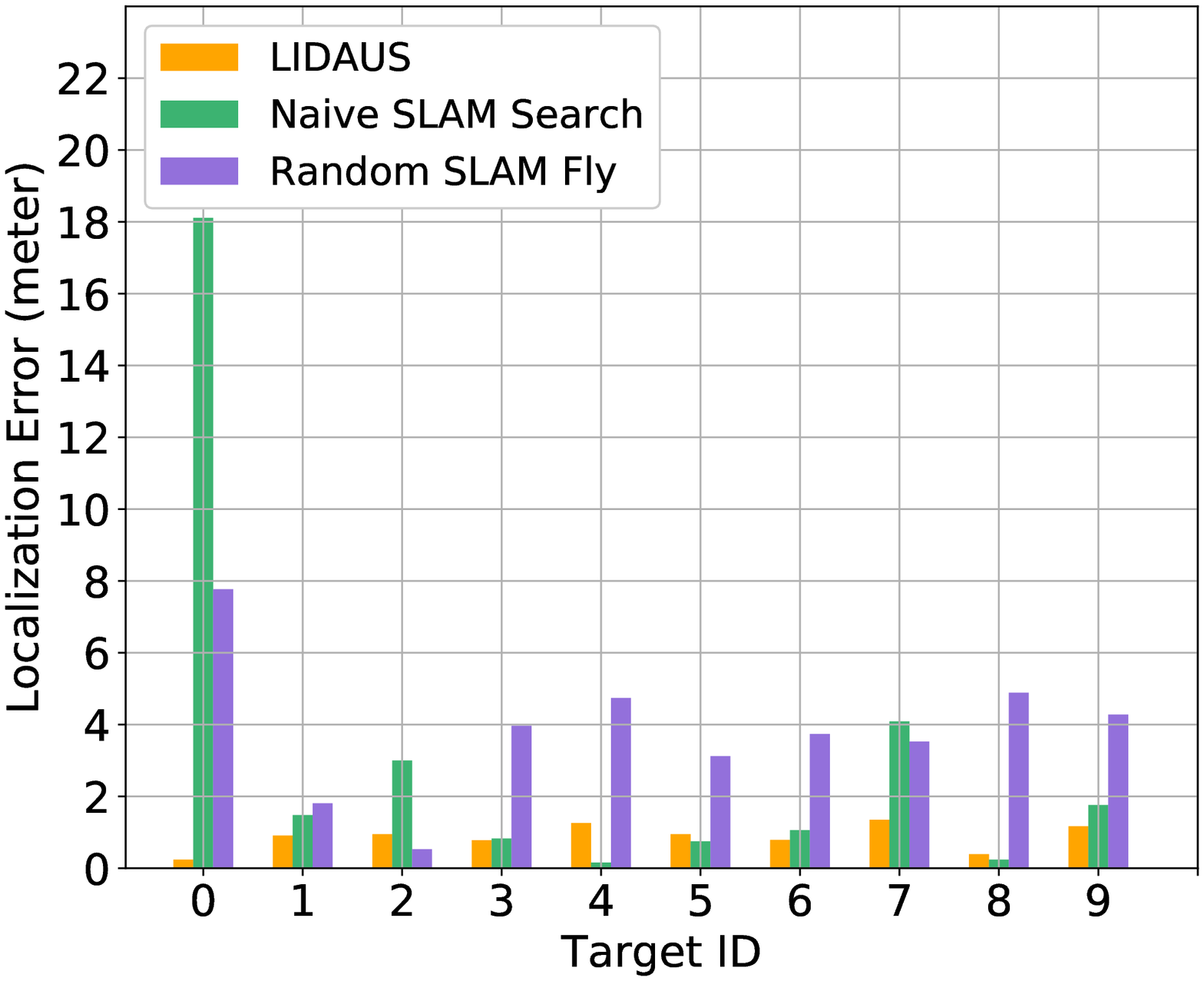}\\
				{}
			\end{center}
			\caption{Bar charts of all 10 targets' localization errors by the three methods.}
			\label{error_bar}
		\end{minipage}
		\begin{minipage}{1.8in}
			\begin{center}
				\setlength{\epsfxsize}{1.8in}
				\epsffile{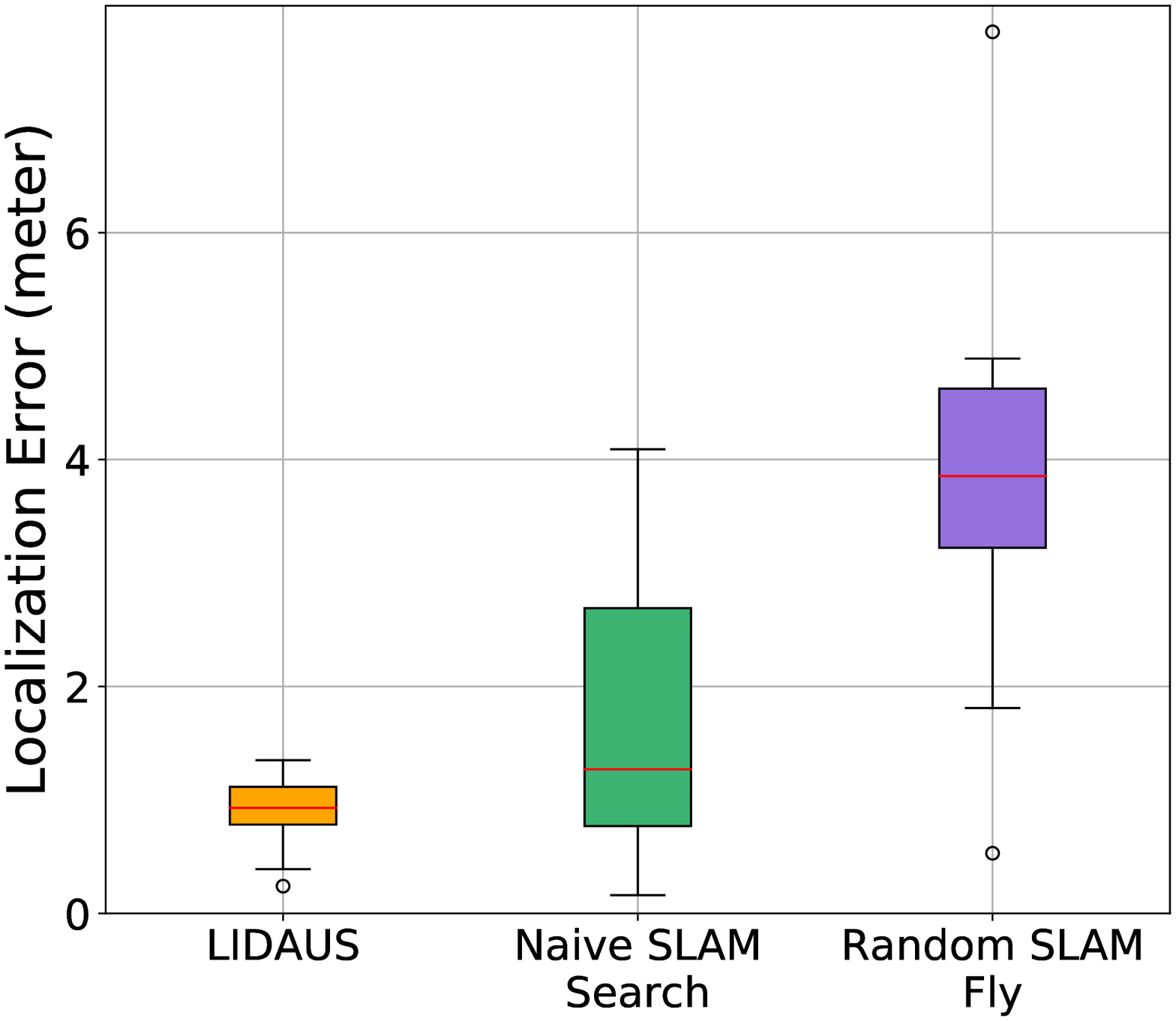}\\
				{}
			\end{center}
			\caption{Boxplots of all 10 targets' localization errors by the three methods. 
				Note that one extreme large error value of Naive SLAM Search is omitted here.}
			\label{error_box}
		\end{minipage}
	}
\end{figure}

We now compare the performance of the three methods. 
Fig. \ref{error_bar} shows the localization errors (i.e., the distance between the true location of a beacon and
its estimated location) of all three methods over all those 10 target beacons. 
We see that 
LIDAUS in general can achieve significantly higher localization accuracy than the other two methods. LIDAUS's 
estimation errors are always no more than $1m$ (except that one beacon's error is slightly higher than $1m$).

\begin{figure}[htb!]
	\centerline{
		\begin{minipage}{2.8in}
			\begin{center}
				\setlength{\epsfxsize}{2.8in}
				\epsffile{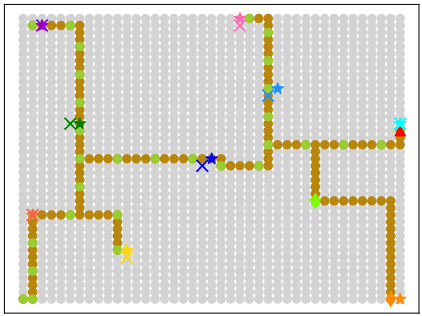}\\
			\end{center}
		\end{minipage}
	}
	\caption{Steiner trees built by LIDAUS in its final searching stage. The red triangle denotes the current position of UAV. Star symbols denote the real locations of target beacons. 
		Diamond symbols denote the current estimated locations of the not-yet-found target beacons. 
		Cross symbols denote the final estimated locations of found target beacons. 
		Light green dots on the branch of the tree denote deployed anchor beacons.}
	\label{trees}	
\end{figure}

\begin{figure}[htb!]
	\centerline{
		\begin{minipage}{2.2in}
			\begin{center}
				\setlength{\epsfxsize}{2.2in}
				\epsffile{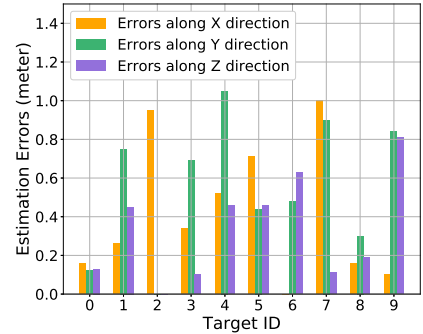}\\
				{}
			\end{center}
		\end{minipage}
	}
			\caption{LIDAUS's localization errors of target beacons along x, y and z directions.} 
			\label{our_xyz_error}
\end{figure}

Fig. \ref{error_box} shows the boxplots of localization errors of the three methods. 
We observe that the average localization errors of LIDAUS, Naive SLAM Search, and Random SLAM Fly are $0.88$, $3.15$ and $3.84$ meters respectively.
LIDAUS's average performance is significantly better than the other two methods. 
Fig. \ref{our_xyz_error} illustrates LIDAUS's localization errors along x, y and z directions. 
We can see that, except that target beacon $4$ has a slightly large error along y-direction ($1.05$ meters), 
all other targets' localization errors along these three directions are limited within $1$ meter. 
As a comparison, neither the other two methods can achieve this level of accuracy. 
Overall, we see that LIDAUS has the lowest average localization error and the lowest localization variance, 
and hence has the best performance.
To further illustrate LIDAUS, Fig. \ref{trees} shows the Steiner tree built on the ground layer in the final searching stage of the UAV, and the locations of anchor beacons. 
During the searching stage, the UAV flies along the shortest branch of the Steiner tree built upon the estimated locations of 
target beacons and deploys anchors every $4m$ on its path. 

\subsection{Experiments}\label{sec_exp}

We now demonstrate our system design via a real-world experiment. 
We built a UAV with Bluetooth receiver module based on Crazyflie \cite{crazyflie}, and we used it in our experiments. 

\begin{figure}[htb!]
	\centerline{
		\begin{minipage}{2in}
			\begin{center}
				\setlength{\epsfxsize}{2in}
				\epsffile{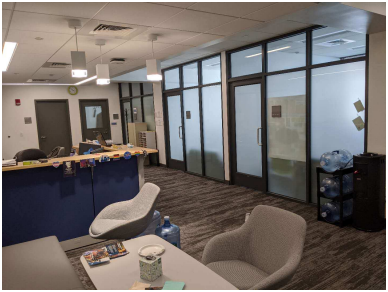}\\
				{}
			\end{center}
		\end{minipage}
	}
	\caption{The office space where experiments were performed.}
	\label{engin_office}
\end{figure}

Eight IoT beacons (i.e., targets) were placed on the ground of an office space as shown in Fig. \ref{engin_office}. 
All targets and anchors (i.e., anchor beacons) advertised at a rate of $2$Hz with a transmission power set to $0$ dBm. 
The UAV was initially placed at the origin (i.e., the main entrance).

Initially the UAV took a Eulerian cycle path to explore the whole space. Due to furnitures in the space,
the grid graph derived by our space discretization contains various sized grid squares or rectangles, and 
most of the grid squares have a side length of $2m$.
Every $50$ cm along its flying path, the UAV collected $20$ sets of RSSI data from beacons in the space. 
During the exploring stage, no anchor was deployed. 
Then it took three \textit{searching stages} to finish its localization of all target beacons.   
In each searching stage, the UAV followed the shortest branch of the Steiner tree built in that stage
when searching for target beacons, and it deployed anchors every $2m$.
In addition, when it reached a target's estimated location as planned, 
it flied along an extra square path ($1m$ side length) with the estimated target position 
as center in order to receive stronger RSSI data for the target.
In addition, we also performed experiments with FastSLAM 1.0 and FastSLAM 2.0, 
following the same flying path as mentioned above.

\begin{figure}[htb!]
	\centerline{
		\begin{minipage}{2in}
			\begin{center}
				\setlength{\epsfxsize}{2in}
				\epsffile{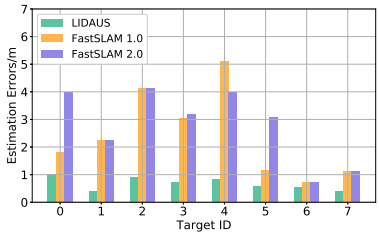}\\
				{}
			\end{center}
		\end{minipage}
	}
	\caption{Our system's average localization error is $0.68m$ over all target beacons. 
		For those estimated locations that were out of the range of the space (when using FastSLAM 1.0 or 2.0), 
		they were rounded to the nearest position in the space.}
	\label{exp_error}
\end{figure}

Fig. \ref{exp_error} shows that our system's localization estimation error for each target 
is no more than $1$ meter, with an average $0.68m$. 
This experiment result shows that our system can achieve a sub-meter localization accuracy. 
Note that both FastSLAM 1.0 and 2.0 performed very poorly. In addition, FastSLAM 2.0 is even worse
in most cases than FastSLAM 1.0, due to the high noise level of sensor data (i.e., RSSI).


\section{Conclusion and Future Work}\label{sec_conc}

We have developed LIDAUS, an infrastructure-free, multi-stage 
SLAM system that utilizes a UAV to search and accurately localize IoT devices
in a 3D space. The system is based only on the RSSIs 
of the existing commodity IoT devices. It can be easily deployed without any customized signal processing hardware
and without requiring any fingerprinting or pre-trained model. It can deal with dynamic changing environment. 
Furthermore, it can work in a complex indoor space where GPS is not available or a space where 
it is inaccessible for humans. The system contains several novel techniques such as a weighted entropy-based clustering
of RSSI observation locations, 3D U-SLAM (and its selective replay) with dynamic deployed anchors, and path planning based on edge covering Eulerian cycles and Steiner tree route for cost minimization. Our extensive simulations and real-world experiments 
have demonstrated the system's effectiveness of IoT device localization. In future work, we will 
further improve the software and hardware design and conduct large scale experiments in consideration of the limited battery capacity of UAVs to achieve high energy efficiency. We will also extend the system to 
include a swarm of UAVs with support from edge computing.

\bibliographystyle{IEEEtran}
\bibliography{social_swarms}

\end{document}